\pdfoutput=1
\documentclass[11pt]{article}

\usepackage[]{emnlp2021}

\usepackage{times}
\usepackage{latexsym}
\usepackage{amsmath}
\usepackage{xcolor}
\usepackage{booktabs}
\usepackage{url}
\usepackage{bm}
\usepackage{array}
\usepackage{graphicx}
\usepackage{graphics}

\usepackage{arabtex}
\usepackage[T1]{fontenc}
\newcommand{\hide}[1]{}

\newcommand{\AHAMZAUP}{{\^{A}}}

\newcommand{\TAMARBUTA}{{$\hbar$}}

\newcommand{\areta}{{\sc Areta}}

\title{Automatic Error Type Annotation for Arabic}

\author{Riadh Belkebir \and Nizar Habash \\
  Computational Approaches to Modeling Language (CAMeL) Lab\\
  New York University Abu Dhabi \\
  \texttt{\{rb4822,nizar.habash\}@nyu.edu} 
  }

\begin{document}
\maketitle
\begin{abstract}
We present {\areta}, an automatic error type annotation system for Modern Standard Arabic.  We design {\areta} to address Arabic's morphological richness and orthographic ambiguity. 
We base our error taxonomy on the Arabic Learner Corpus (ALC) Error Tagset with some modifications. {\areta} achieves a performance of  85.8\% (micro average F1 score)
on a manually annotated blind test portion of ALC.
We also demonstrate {\areta}'s usability by applying it to a number of submissions from the QALB 2014 shared task for Arabic grammatical error correction.  The resulting analyses give helpful insights on the strengths and weaknesses of different submissions, which is more useful than the opaque M$^2$ scoring metrics used in the shared task.
{\areta} employs a large Arabic morphological analyzer, but is completely unsupervised otherwise.
We make {\areta} publicly available.

\end{list}
\end{abstract}
\setarab
\novocalize

\section{Introduction}
%\cite{alfaifi2014evaluation}
%\cite{alfaifi2015building}

There has been a lot of interest recently in Automatic Error Evaluation for many languages. Many specialized shared tasks in grammatical error correction (GEC) and text normalization have used tools like M$^2$ Scorer \cite{dahlmeier-ng-2012-better} and ERRANT \cite{bryant2017automatic}.   
In contrast with the opaque results of M$^2$ Scorer based on extracted edits, ERRANT, designed primarily for English, allows for deep interpretation of GEC error types since it gives more detailed explanations.  Error type explainability is helpful for many NLP applications, including second language learning.

\begin{figure}[t!]
    \begin{centering} 
   \includegraphics[width=0.26\textwidth]{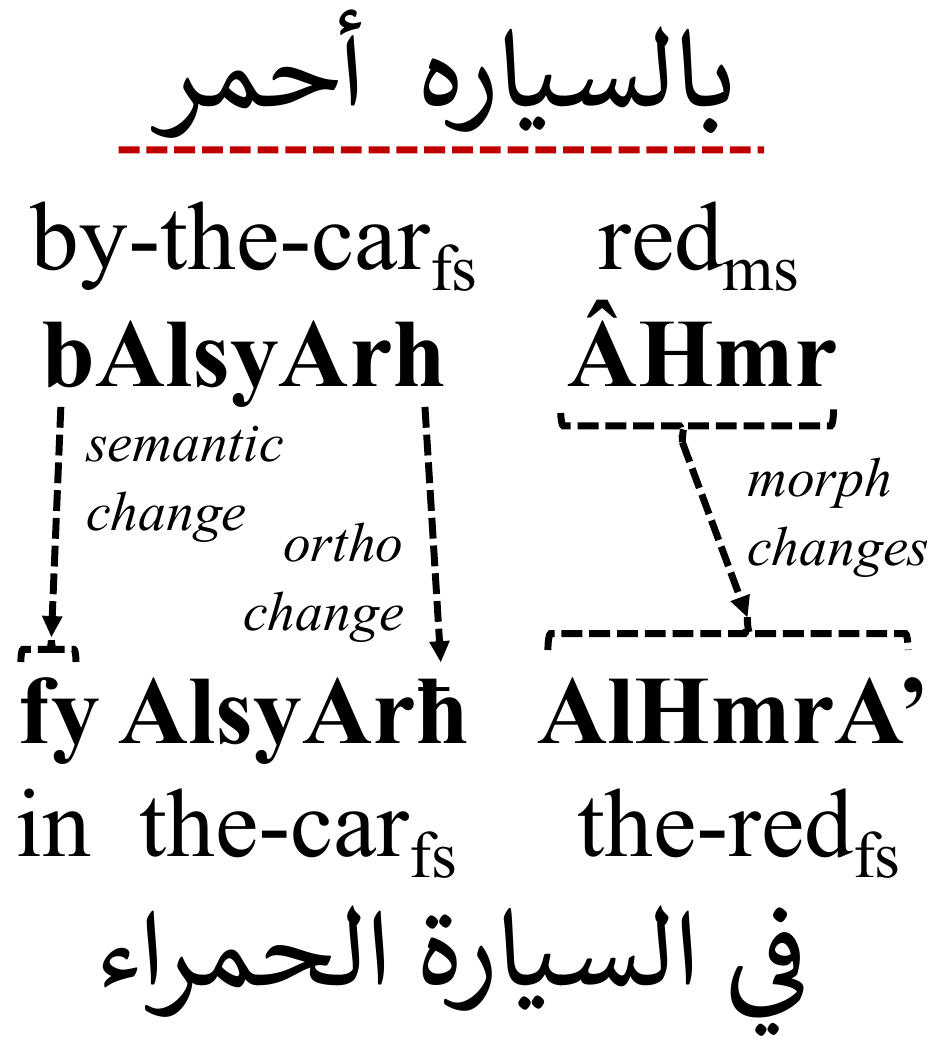}
    \caption{An example of aligned erroneous source and reference phrases with three different error types.}
    \label{fig:example}
    \end{centering}
\end{figure}

Arabic is a morphologically rich and complex language with a high degree of ambiguity at the orthographic, morphological,  syntactic, lexical and semantic levels \cite{Habash:2010:introduction}. 
Figure~\ref{fig:example} presents a motivating example for the complexity of Arabic  error type annotation (discussion in Section~\ref{sec:example}).
Previous Arabic text correction shared tasks like QALB 2014 \cite{mohit-etal-2014-first} and QALB 2015 \cite{rozovskaya-etal-2015-second} evaluated submissions using the M$^2$ Scorer.
\newcite{alfaifi2015building} proposed a taxonomy for Arabic error types and annotated the Arabic Learner Corpus (ALC) using it; but, he does not provide an error classification tool.

In this paper, we present {\areta}, a system for the extraction and annotation of error types in Arabic.  {\areta} is inspired by ERRANT, but addresses the unique and complex challenges of Arabic. We base our error taxonomy on the ALC Error Tagset \cite{alfaifi2014evaluation,alfaifi2015building} with some modifications. {\areta} reaches a micro average F1 score of 85.8\% on an ALC blind test.
We also demonstrate {\areta}'s usability on a number of submissions from QALB 2014 \cite{mohit-etal-2014-first} shared task.
While {\areta} employs a large Arabic morphological analyzer, it is completely unsupervised otherwise.
To our knowledge, this is the first system of its kind for  Arabic. 
We make {\areta} publicly available.\footnote{\url{https://github.com/CAMeL-Lab/arabic_error_type_annotation}}

The remainder of this paper is organized as follows. Sections~\ref{sec:example} and~\ref{sec:rw} present a motivating example, and  related work, respectively. Section~\ref{sec:approach} describes our approach. Section~\ref{sec:exp}  presents experimental results and discussions.
%the strengths and limitations of the system. 
%In section 7, we present the conclusion. 

%%%%%%%%%%%%%%%%%%%%%%%%%%%%%%%%
\section{A Motivating Example}
\label{sec:example}
Figure~\ref{fig:example} shows an example of an erroneous Arabic source phrase <bAlsyArh 'a.hmr>  {\it bAlsyArh {\AHAMZAUP}Hmr}\footnote{All Arabic script examples are paired with a strict 1-to-1 transliteration in the 
HSB scheme \cite{Habash:2007:arabic-transliteration}.} and its correct reference <fy AlsyArT Al.hmrA'>  {\it fy AlsyAr{\TAMARBUTA} AlHmrA'}.  The phrase, meaning `in the red car', includes three  error types. 
%: orthographic, morphological, and semantic.
%First, the two sentences are aligned where we have the following pairs (<b>~b~$\rightarrow$~<fy>~fy), (<AlsyArh>~{\it AlsyArh}~$\rightarrow$~<AlsyArT>~{\it AlsyArp}) and (<A.hmr>~{\it AHmr}~$\rightarrow$~<al.hmrA'>~{\it AlHmrA'}). In a second step, we feed the different pairs to the explainability model that will give tags to the different pairs of the two sentences as follow:
\begin{itemize}
    %\setlength\itemsep{0.1em}
    %Add Buckwalter transliteration citation.
    \item  (+<b>~{\it b+}~$\rightarrow$~<fy>~{\it fy})\\{\bf Semantic error}: the preposition proclitic +<b>~{\it b+} `by/with' is used instead of the free preposition <fy>~{\it fy} `in'.
    \item (<AlsyArh>~{\it AlsyArh}~$\rightarrow$~<AlsyArT>~{\it AlsyAr{\TAMARBUTA}})\\
    {\bf Orthographic error}: the Ta-Marbuta feminine marker, <T> {\it {\TAMARBUTA}}, is misspelled as <h> {\it h}.
    \item (<'a.hmr>~{\it {\AHAMZAUP}Hmr}~$\rightarrow$~<al.hmrA'>~{\it AlHmrA'})\\ {\bf Morphological errors}: (a) masculine gender is used instead of feminine, and (b)  the definite article proclitic +<al>~{\it Al+} `the' is dropped.
\end{itemize}

A simple Levenshtein edit distance \cite{Levenshtein:1966:binary} between the source and reference phrases suggests the reference is modified through two word substitutions and one word deletion, or three character substitutions and five character deletions. In contrast, a linguistically motivated error type classification is more insightful.  

From the point of view of the source phrase, there are two words, and they each get two error tags according to the ALC error taxonomy (Table~\ref{tab:taxonomy}). The first word <bAlsyArh>~{\it bAlsyArh} has an attachable proclitic and as such includes both semantic and orthographic errors. And the second word <'a.hmr>~{\it {\AHAMZAUP}Hmr} includes two morphological errors (gender and definiteness).
A system to identify the exact error types needs to be aware of not only the complexity of Arabic morphology but also the possibility of multiple co-occurring error types. We address  these issues in {\areta}'s design.

%Also, we give special attention to the most common errors in Arabic text correction at the orthographic level, such as Ha and Ta-Marbuta errors. Our framework is completely unsupervised and gives error type explainability at different granularity levels, which we will discuss in the taxonomy subsection.

%%%%%%%%%%%%%%%%%%%%%%%%%%%%%%%%

\begin{table*}[t!]
    \centering{
   \includegraphics[width=0.97\textwidth]{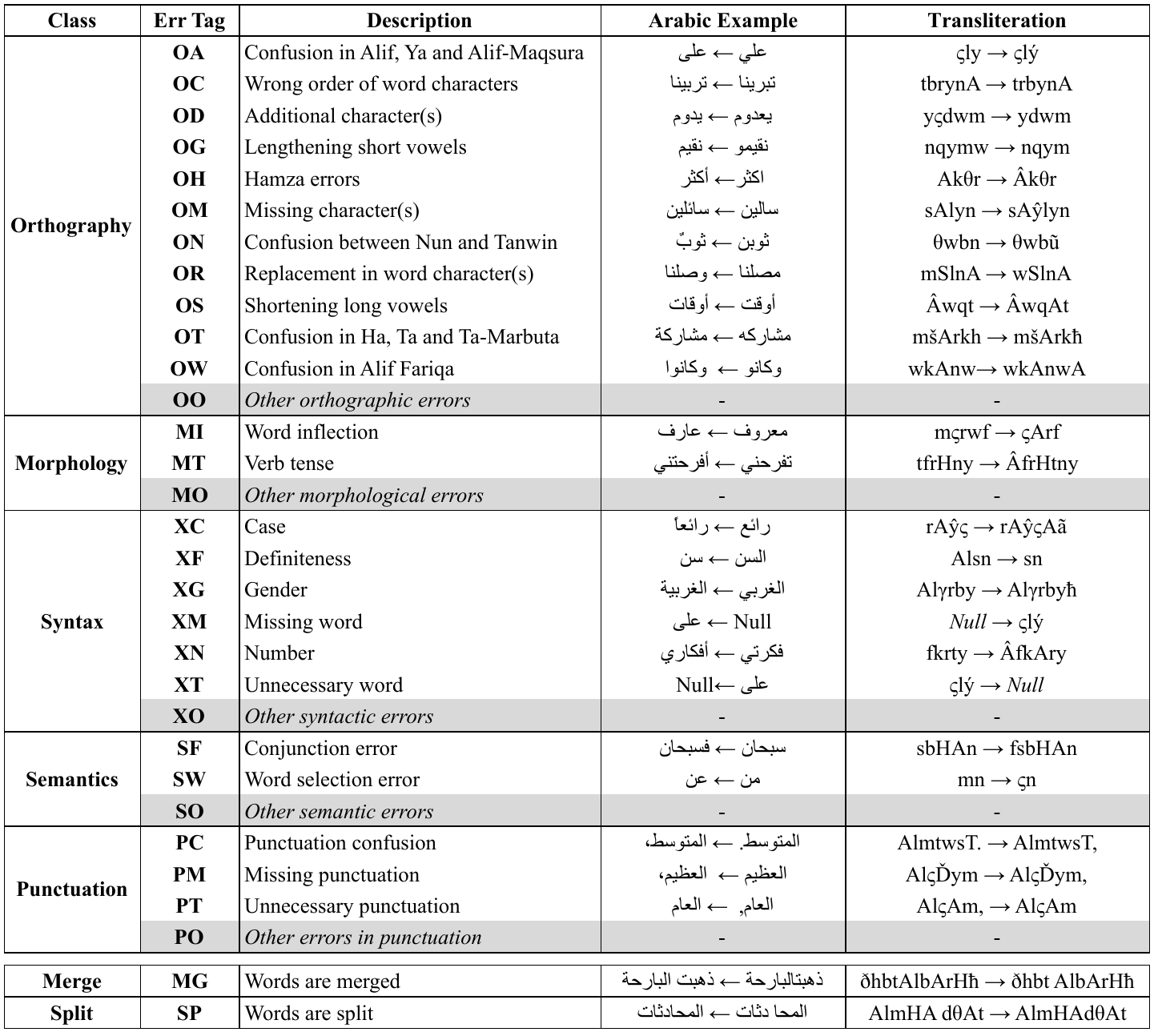}
    \caption{The ALC error type taxonomy extended with merge and split classes. The error tags are listed alphabetically, except for the highlighted {\it *Other} tags, which we do not support in {\areta}.}
    \label{tab:taxonomy}
    }
\end{table*}

\section{Related Work}
\label{sec:rw}
While M$\textsuperscript{2}$ Scorer \cite{dahlmeier-ng-2012-better} has been used for automatic evaluation of GEC shared tasks in different languages \cite{ng-etal-2014-conll,ng2013conll,mohit-etal-2014-first,rozovskaya-etal-2015-second},
a lot of attention has been paid to annotating and evaluating the output of English text correction systems, e.g., ERRANT \cite{bryant2017automatic}.   There is still a lack of tools that allow such utility for other languages, including Arabic. In the rest of this section, we present the main tools for evaluating and annotating error types, some of the challenges of Arabic processing, and the Arabic error taxonomy which we modify.

\subsection{M$\textsuperscript{2}$ Scorer, ERRANT, and SERRANT}
%\paragraph{M$\textsuperscript{2}$ Scorer}
The M$\textsuperscript{2}$ Scorer \cite{dahlmeier-ng-2012-better} is a tool used for evaluating GEC systems based on F1 or F0.5 scores. It uses a method called $MaxMatch$ (M$^2$) to compute the sequence of phrase-level edits that achieve the highest overlap with the gold (reference) annotation. Although the M$\textsuperscript{2}$ Scorer evaluates  GEC systems based on  extracted edits, it does not provide  error types based on the linguistic features of the language.

ERRANT \cite{bryant2017automatic} is a grammatical ERRor ANnotation Toolkit that automatically extracts edits from parallel original and corrected sentences and classifies them using a rule-based framework. It was first applied to the CoNLL-2014 shared task \cite{ng-etal-2014-conll} to carry out detailed error type analyses. Most current GEC systems use ERRANT to annotate extracted edits and evaluate system outputs. The ERRANT taxonomy has 25 main error categories.
%\subsection{SERRANT}

SERRANT \cite{choshen2021serrant} is a system for automatic classification of English grammatical errors that combines ERRANT \cite{bryant2017automatic} with {\sc SErCl}, a taxonomy of Syntactic
Errors and an automatic Classification \cite{choshen-etal-2020-classifying}. SERRANT uses ERRANT's annotations when they are informative and those provided by {\sc SErCl} otherwise.

While the M$\textsuperscript{2}$ Scorer is generic and can be applied to many languages to extract edits and evaluate GEC system quality, ERRANT and SERRANT focus more on linguistic aspects and give better explainability of error types. However, these frameworks require knowledge about the targeted language and are expensive to build. Furthermore, the ambiguity challenges that are part of the Arabic language make the task even more challenging since the error types can be interpreted differently for many words.

\subsection{Arabic Language Processing Challenges}
%cite habash book in final
Arabic poses a number of challenges for natural language processing 
in general and the task of grammatical error correction and error type annotation in particular \cite{Habash:2010:introduction}.
First, Arabic is morphologically rich. Words in Arabic inflect for person (per), gender (gen), number (num), aspect (asp), voice (vox), mood (mod), state (stt) and case (cas). Furthermore, Arabic uses a number of attachable proclitics (prc0-2) and enclitics (enc0). 
Second, Arabic is orthographically very ambiguous due to the use of optional diacritics, which are almost always absent.  Figure~\ref{tab:morph-example} demonstrates the various analyzes associated with two Arabic words. In some cases the analyses differ in part of speech (POS).  
To address these challenges, {\areta} uses CAMeL~Tools \cite{obeid:2020:cameltools}, an open source Python toolkit for Arabic language processing. CAMeL~Tools uses the CALIMA-Star Arabic morphological analyzer \cite{taji-etal-2018-arabic} and provides morphological disambiguation functionality over it.  
%We use the light maximum likelihood estimation (MLE) model as part of {\areta}.
%CALIMA-Star extends upon the Standard Arabic Morphological Analyzer \cite{Graff:2009:standard}.

In developing {\areta}, we took inspiration from AMEANA \cite{ElKholy:2011:automatic}, which also relies on morphological analyzers to provide morphological error analysis in the context of machine translation evaluation.  However, {\areta} addresses a wider range of error types, and is intended to be more general.

\subsection{The Arabic Learner Corpus Error Taxonomy}
 \newcite{alfaifi2014evaluation} proposed a taxonomy of 29 error tags for Arabic (See Table~\ref{tab:taxonomy}). They annotated a portion of Arabic Learner Corpus (ALC) dataset
% The full ALC includes 282,732 words
%- 1585 materials (written and spoken)
%- Produced by 942 students
%- from 66 different L1 backgrounds
%- Studying at 25 Educational institutions
%
such that for each erroneous word, one of the classes of error is given along with the word's correction. In \newcite{alfaifi2015computer} they presented a tool that facilitated {\it semi-automatic error tagging}. The tagging feature worked as translation memory, where annotated words are saved in a database, and recalled when seen again. 

%The work presented in  \newcite{alfaifi2015computer} allows a better understanding of the error types of the Arabic language. However, the semi-automatic way of the annotation process is rather limited. Therefore it is not possible to evaluate Arabic GEC systems automatically.
%\subsection{Taxonomy and automatic error type annotation} 
%\subsubsection{Taxonomy}
%Based on our review of previous works on Arabic error type annotation, we found that \cite{alfaifi2014evaluation} is the one that is available and has a comprehensive taxonomy. They also provided annotated data which allows us to evaluate the annotation process of our framework. Therefore, we adapted this taxonomy and suggested some modifications to it.
We base the error taxonomy we use in {\areta} on \newcite{alfaifi2014evaluation}'s comprehensive taxonomy with two modifications. First, we add two error classes - merge (MG) and split (SP) to allow handling man-to-many word corrections. And secondly we drop all of the {\it Other} error tags - OO, MO, XO, SO and PO, corresponding to {\it other} orthographic, morphological, syntactic, semantic and punctuation errors, respectively.  These errors tags collectively accounted for 0.7\% of all error tags in the ALC, and would have added a lot of complexity to our system.
As such, {\areta}'s full taxonomy has 7 classes and 26 error tags.  When we evaluate {\areta} against the ALC annotations, we penalize {\areta} for missing all the {\it Other} tags.

\section{Approach}
\label{sec:approach}
%\section{Problem Statement and Research Questions}
%\label{sec:ps}
In this section, we present our approach to developing {\areta},
an automatic error annotation framework for Arabic. 
We organize this section in three parts: basic word alignment, automatic error annotation,  and error-type-based evaluation.

Given a raw input sequence ($S_{raw}$), a system output sequence (hypothesis) ($S_{hyp}$), and a reference sequence ($S_{ref}$), we want to be able to annotate and evaluate the quality of the system output (hypothesis) ($S_{hyp}$).

%In the following subsections, we present the methodology we followed to build the framework that aligns, annotates, and evaluates the Arabic GEC systems.
%Based on our review of previous related work on GEC evaluation, we identified a number of research gaps, among them the lack of error annotation and evaluation frameworks for low-resource languages such as Arabic. 

%Therefore, in this paper, we suggest an error annotation and evaluation framework for Arabic and try to address the following research questions:

%\begin{itemize}
%\setlength\itemsep{0.05em}
%\item{\textbf{RQ1:}} What is the taxonomy of error types that we should consider for Arabic GEC?
%\item{\textbf{RQ2:}} How do we align and extract the edits?
%\item{\textbf{RQ3:}} What is the best choice between a morphological analyzer and a disambiguator for Arabic error type annotation?
%\item{\textbf{RQ4:}} What is the evaluation methodology that we should consider?
%\end{itemize}
%
%\section{Method}

\begin{figure*}[t!]
    \centering
   \includegraphics[width=0.94
   \textwidth]{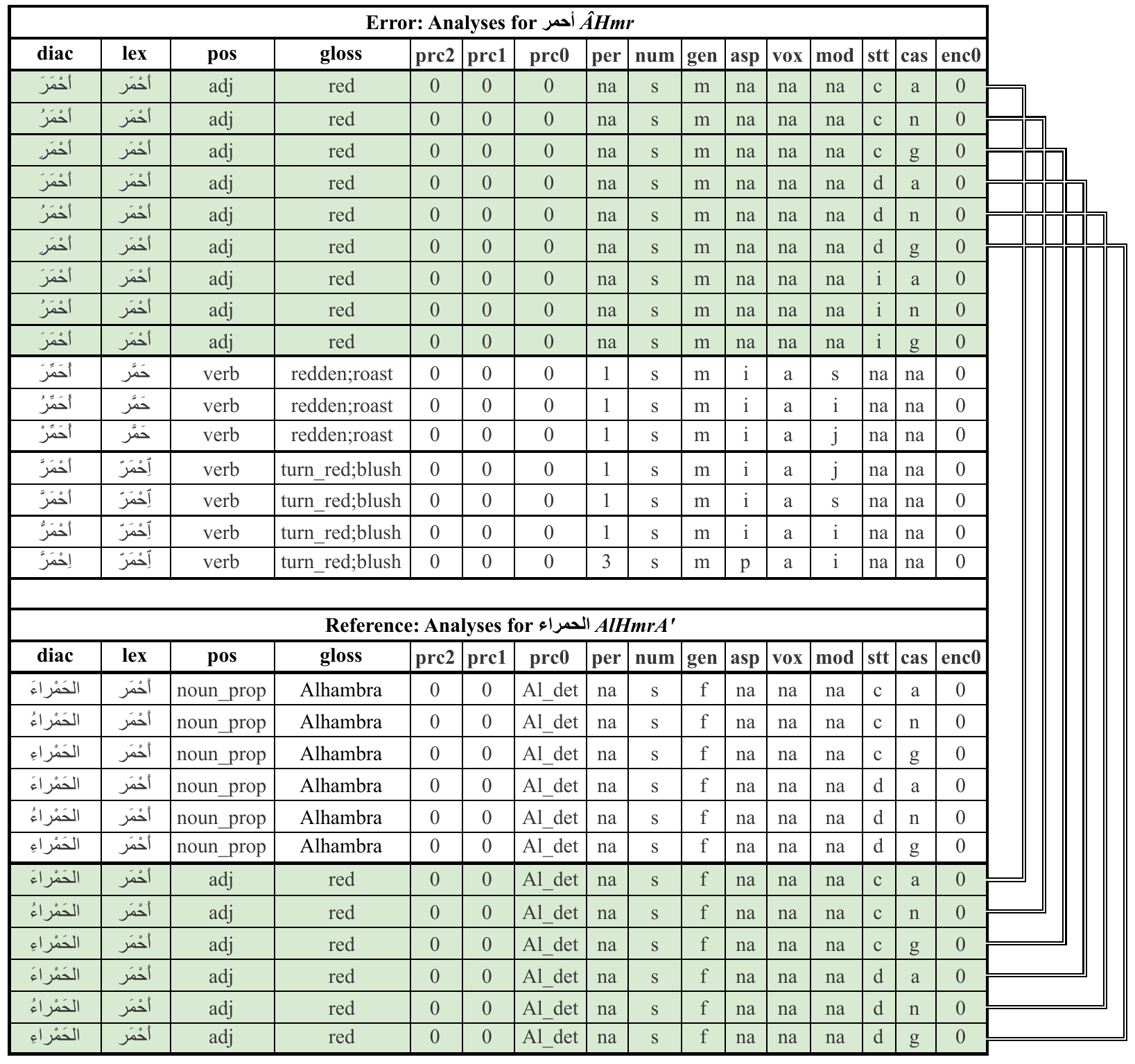}
    \caption{Morphological changes between the words
    <'a.hmr>~{\it {\AHAMZAUP}Hmr} `red$_{ms}$' and <al.hmrA'>~{\it AlHmrA'} `the red$_{fs}$'}
    \label{tab:morph-example}
    
\end{figure*}

\subsection{Basic Word Alignment} 
\label{alignment}
Before we can annotate a word's error type, we need to align said word to its correction.
The first step is thus to word-align the two sequences (source and target) whose differences we want to annotate. These may be the pair of 
$S_{raw}$ and $S_{ref}$, $S_{raw}$ and $S_{hyp}$, or $S_{hyp}$ and $S_{ref}$.
Since this task assumes the source and target to be of the same language with some differences in spelling, it is a simpler task than general word alignment \cite{Och:2003:systematic}.
We start with character-level edit-based alignment to align the characters, and then we group them in words such that the source is aligned to target as 1-to-many words (where many include zero).  We make extensive use of the CED word alignment tool by \newcite{Khalifa:2021:Character}. We refer in the next step to the generated word alignment using the nomenclature of $Align_{source, target}$, e.g., $Align_{raw, ref}$.
%
%To validate the correlation between our alignment and ALC's manual alignment, we tried to predict the error types on both modes (auto vs. manual alignment). According to the results, we find a 99\% correlation.

One limitation of the current implementation of this step is that it cannot handle many-to-many alignments; and as such split errors cannot be modeled in {\areta} unless they are independently provided.  The ALC data does not annotate split nor merge errors, so this limitation has no effect on it.
However, when working with the QALB 2014 data, we exploited the shared task's {\tt .m2} file information which provided some of the merge alignment for raw and reference. These files were created off the  QAWI interface \cite{Obeid:2013:web-based} which was used in the QALB project annotation \cite{Zaghouani:2014:large}.  

We evaluate our word alignment against the manual alignment in the ALC corpus' raw and reference sequences.  Our automatic word alignment matches the manual alignment for 99.24\% of the words.  
The failed alignment cases include minor word reordering 
such as <lm 'akn> {\it lm {\AHAMZAUP}kn} `not-past I-be' aligning with <knt lA> {\it knt lA}  `I-was not', where the negation particles appear at either side of the verb. Other failed alignment cases include inserted words that could be paired with their left or the right neighbors. For example, in the raw sequence <wkAn kl> {\it wkAn kl} `and-it-was all' which is paired with reference <wknt fy kl> {\it wknt fy kl} `and-I-was in all', the word <fy> {\it fy} is paired as <wknt> {\it wknt} in the manual alignment, but the automatic aligner pairs it with <kl> {\it kl}. Both alignments are plausible.

\subsection{Automatic Error Type Annotation}
%The automatic error annotation process takes as an input $Align_{raw, ref}$ and $Align_{raw, hyp}$. 

The input to the automatic error annotation process is an alignment, e.g., $Align_{raw, ref}$. 
Then for each aligned pair of words, the system tries to extract the error type. The system is divided into four components to allow modeling combinations of error tags, but with some constraints driven by efficiency and control over the error tag search space.

First, the {\bf Punctuation component} detects all punctuation error tags (PC, PM, and PT) using regular expressions. The punctuation error tags can be used with tags detected by other components.

Second, the {\bf Regex component} uses regular expressions to detect all merges (MG) and splits (SP), word insertions (XT) and deletions (XM), as well as some orthographic errors (OC, OG, ON, OS, and OW). In the current implementation, this component is terminal if an error tag is identified. Otherwise, we move to the third component if we have a 1-to-1 word pairing, or to the fourth component if we have a 1-to-many word pairing. Many-to-many word pairings are not supported in this version.

Third, the {\bf Ortho-Morph component} handles the more challenging orthographic and morphological  error types and their combination. This component relies heavily on the CAMeL~Tools'  morphological analyzer to handle Arabic's rich morphology and ambiguous orthography~\cite{obeid:2020:cameltools}. The algorithm is as follows:
%It uses a  morphological analyzer (full analyses), or an MLE disambiguator from CamelTools \cite{obeid:2020:cameltools} as two possible options.

For each {\it $Pair_{i}=<raw\_word{_i},ref\_word{_i}>$} in $Align_{raw, ref}$:

%Let's consider that we have as input $Align_{raw, ref}$ and we aim to generate the annotation for a given pair 
%{\it $Pair_{i}$ = $<$$raw\_word{_i}$, $ref\_word{_i}$$>$} of this list of alignments. We proceed as follow:

\begin{enumerate}
    \setlength\itemsep{0.05em}
    \item Generate the list of possible orthographic edits ($edit\_list$) that transforms $raw\_word{_i}$ to $ref\_word{_i}$.
    \item Generate the possible subsets of elements of $edit\_list$ including the empty set. We call this list $p\_edits$.
    \item For each subset $p\_ed_k$ of $p\_edits$, calculate the morphological feature differences ($morph\_edits$) between $raw\_word{_i}$ and $ref\_word{_i}$ after applying $p\_ed_k$ to $raw\_word{_i}$.\footnote{$morph\_edits$ are calculated over analyses of  $raw\_word{_i}$ and $ref\_word{_i}$ that share the same lemma ($lex$) and $pos$ tags.}
    This generates a path of edits $path =[orth\_edits+morph\_edits]$.  
    \item Add $path$  to the list of possible paths ($paths$).
    \item Return the shortest path from the list of possible paths ($paths$).
\end{enumerate}

%The second step of the algorithm is a bit challenging, knowing the high level of ambiguity for the Arabic language. 
Figure \ref{tab:morph-example} demonstrates the step of identifying $morph\_edits$ between  
$raw\_word$ <'a.hmr>~{\it {\AHAMZAUP}Hmr} `red$_{ms}$' and $ref\_word$ <al.hmrA'>~{\it AlHmrA'} `the red$_{fs}$'. The green-shaded analyses represent the subset of all analyses sharing the same lemma and POS. The six linked pairs of analyses all have the same smallest number of $morph\_edits$ (2):  
$prc0:0~\rightarrow Al\_det$  and $gen:m \rightarrow f$. 

%In Figure \ref{tab:morph-example}, we illustrate how we capture the feature changes between the erroneous word <'a.hmr>~{\it {\AHAMZAUP}Hmr} `red$_{ms}$' and   the correct word <al.hmrA'>~{\it AlHmrA'} `the red$_{fs}$'. 
%First, we use Camel Tools~\cite{obeid:2020:cameltools} morphological analyzer to generate the analyses of both words. 
%Then, we match the analyses that share the same lemma ($lex$) and $pos$ tags, which are highlighted in the green area. This will result in multiple possible options for a change (added to the possible $paths$ in step 4). In the end, we return the path with the minimum number of changes. In this case, it will be two changes. 
%The first change consists in adding a determiner $prc0$ (0 ~$\rightarrow$~ Al\_det)  and the second is gender change $gen$ (m ~$\rightarrow$~ f). 

The final step in the third component uses rules to map the set of edits in the shortest edit path to corresponding  error tags.  For example, any  $morph\_edit$ involving $gen$ will result in the $XG$ error tag, and any $orth\_edit$ involving a Ta-Marbuta change results in the $OT$ error tag.  As such, the example in Figure \ref{tab:morph-example} receives the complex error tag $XF$+$XG$ (definiteness and gender).  
In cases with only $orth\_edit$, we map to  additional/missing/replaced character error tags ($OD$, $OM$, $OR$) when the percentage of affected raw word characters is below 50\%. In cases above that heuristic threshold, we assign the word selection error tag ($SW$).  

Fourth, the {\bf Multi-Word component} handles 1-to-many
word pairings by applying Arabic Treebank (ATB) tokenization to both sides \cite{Maamouri:2004:developing,Habash:2010:introduction}.  ATB tokenization splits all clitics except for the definite article.  We generate the unique ATB tokenizations for all the possible analyses using the CAMeL~Tools morphological analyzer \cite{obeid:2020:cameltools}.  At this point, for each tokenized sequence pair (e.g., raw and reference), we apply the basic word alignment step (Section~\ref{alignment}) locally, and pass the aligned ATB tokens through components one, two and three. The resulting error tags for the various ATB tokens are joined and assigned to the word that produced them.
In the example in Figure~\ref{fig:example}, the 1-to-many pairing of <bAlsyArh> {\it bAlsyArh} and <fy AlsyArT> {\it fy AlsyAr{\TAMARBUTA}} is handled by this component, and receives the complex error tag $SW$+$OT$ (word selection and Ta-Marbuta).

%So for the example provide in figure \ref{fig:example}, the tags that will be assigned according to Al-Faifi's taxonomy are as follow: \\
%({\it b} ~$\rightarrow$~ {\it fy}): $SW$,\\ 
%({\it AlsyArh} ~$\rightarrow$~ {\it AlsyAr{\TAMARBUTA}}): $OT$,\\
%and ({\it {\AHAMZAUP}Hmr} ~$\rightarrow$~ {\it AlHmrA'}): ($XG$+$XF$).

%All words with corrections, that are not assigned any error tags by the system components, are given the UNK error tag. 

\subsection{Error-Type-based Evaluation}
\label{errortype}
{\areta} can be used to conduct error-type-based evaluations in a number of configurations.

First, given reference error tags, as in triplets of ($S_{raw}$, $S_{ref}$, $ErrorTag$),
we can evaluate how well {\areta} performs in automatic error-type annotation in terms of 
F1-score (Micro Avg, Macro Avg, and Weighted Avg) of the different error tags.
See Section~\ref{sec:ALC}.

Second, in the case of no reference error tags, we use our system to identify the {\it reference} error tags using the pair ($S_{raw}$, $S_{ref}$) and compare them using F1 score to the predicted error tags using the pair ($S_{raw}$, $S_{hyp}$).  See Section~\ref{sec:QALB}

Finally, {\areta} can be also used to diagnose a system's output given the reference directly ($S_{hyp}$, $S_{ref}$) to identify {\it remaining error types}.

%Given the assigned error types to the words in a text, we can evaluate the correctness of the text against an independent error type assignment to the same text using
%In the case where we have gold reference error types, as in the ALC annotations ($S_{raw}$, $S_{ref}$, $ErrorType$), we simply compare our system's proposed error types on the ALC error types (Section~\ref{sec:ALC}).

%The first step is to align the raw sentence ($S_{raw}$) with the reference sentence ($S_{ref}$) and the raw sentence ($S_{raw}$) with the system hypothesis sentence ($S_{hyp}$). The result of the alignment between $S_{raw}$ and $S_{ref}$ will be a list of pairs $Align_{raw, ref}$. The alignment between $S_{raw}$ and $S_{hyp}$ is also a list of pairs $Align_{raw, hyp}$. The size of $Align_{raw, ref}$ is equal to the size of $Align_{raw, hyp}$. This will allow in a further stage to annotate the elements of pairs of  $Align_{raw, ref}$ and $Align_{raw, hyp}$ and also evaluate the performance.

\begin{table*}[t!]
    \centering{
  \includegraphics[width=0.93\textwidth]{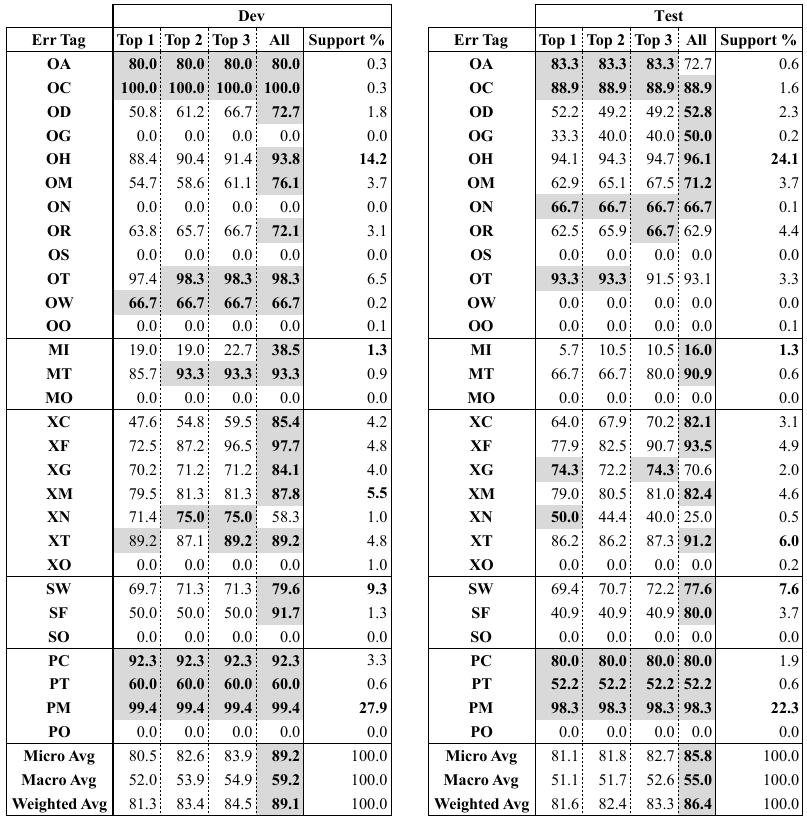}
    \caption{Comparing the F1 score results for error tag annotation on ALC {\bf Dev} and {\bf  Test} sets using different number of analyses from the morphological analyzer.}
    \label{tab:dev+test-mle-vs-analyzer}
    }
\end{table*}

%\begin{table}[t!]
%    \centering{
%  \includegraphics[width=0.37\textwidth]{tables/test_faifi.pdf}
%    \caption{Comparing the results of using CamelTools morphological %analyzer and MLE for error type annotation on the ALC {\bf test} set.}
%    \label{tab:test-mle-vs-analyzer}
%    }
%\end{table}
%

\section{Experiments}
\label{sec:exp}

%To evaluate the framework, we used data provided by \cite{alfaifi2014evaluation}. So after the annotation, we calculate the precision, recall, and F1-score for each error type category. 
We present next two sets of experiments. First,  we evaluate the quality of error type annotation in comparison to the ALC. Second, we calculate the correlation between an evaluation metric based on our error type prediction system and the M$^2$ Scorer.
%, we reevaluate submitted systems from the QALB 2014 \cite{mohit-etal-2014-first} shared task. 

\subsection{Datasets}
To perform the experiments, we used two datasets. 
First is the {\bf Arabic Learner Corpus} \cite{alfaifi2014evaluation}, which contains 10K words annotated for error type and corrections.
The number of changed raw words is 1,688 ($\sim$16.8\%),  of which  75\% appear only one time and  12\% appear two times. Because of this sharp-tailed Zipfian distribution and limited training instances, we expect it to be hard to learn from this data using machine learning systems. 
We split the data into two parts: Dev and Test by randomly selecting complete documents from the corpus (10 for Dev and 10 for Test).\footnote{The Dev set consists of the files with the prefix: {\it S038, S437, S448, S498, S505, S664, S785, S793, S927} and {\it S931}. The Test set consists of the files with the prefix: {\it S037\_(T1|T2), S274, S301, S496, S662, S670, S799}, and {\it S938\_(T1|T2)}.} Dev was used while building {\areta}.

%Among them 146 words have multiple errors which represents 8.6\% of the data.  
The second dataset is the {\bf QALB 2014 Shared Task} test set and some of the submitted systems' outputs \cite{mohit-etal-2014-first}. We use it to evaluate the correlation between the M$^2$ Scorer and our metrics: F1-score (Micro Avg, Macro Avg, and Weighted Avg) of the different error tags.

\subsection{ALC Automatic Error Annotation}
\label{sec:ALC}
Table~\ref{tab:dev+test-mle-vs-analyzer} presents the results of evaluating {\areta}'s performance in identifying error tags on the Dev and Test portions of the ALC data set.

We compare four settings that vary in terms of the number of analyses from the CAMeL~Tools morphological analyzer \cite{obeid:2020:cameltools}: using the top 1, 2, or 3 analyses from the MLE disambiguator, or using all analyses (16 analyses/word).  On average, using the top 1, 2 or 3 analyses took about the same time ($\sim$16 secs to run Dev), while the full analyzer took 44\% more time ($\sim$23 secs to run on Dev).\footnote{2.4 GHz 8-Core Intel Core i9 machine.}
%>>
Consistently, in both Dev and Test, using more analyses improves the performance of {\areta} across all metrics. 
{\areta}'s best setting (All) matches the ALC annotation with 89.2\% (F1 Micro Avg) on Dev, and 85.8\% (F1 Micro Avg) on Test.
For some specific error tags the performance drops with more analyses, due to the larger search space introduced by the analyzer.

We observe that among the top five tags in terms of frequency in Dev, {\areta} detects, with high accuracy, errors involving punctuation ($PM$), Hamzas ($OH$), and Ha/Ta/Ta-Marbuta ($OT$); however the performance on word selection ($SW$) and missing words ($XM$) are lower.  The distribution of error tags varies between Dev and Test sets: there is a 91.3\% correlation between the support of the tags in the two sets, but the top three error tags are the same in both ($PM$, $OH$, and $SW$). The top five tags in terms of frequency  in Test also include unnecessary words $XT$ and definiteness $XF$, both of which perform relatively well.
The F1 Macro Average of the top five error tags (in terms of support) is 91.8\% for Dev, and 91.3\% for Test.

If we group the error tags by the components that handle them, the F1 weighted averages for Punctuation ($PC$, $PM$, $PT$), Regex ($OC$, $OG$, $ON$, $OS$, $OW$, $XM$, $XT$), and  Ortho-Morph components ($MI$, $MT$, $OA$, $OD$, $OH$, $OM$, $OR$, $OT$, $SF$, $SW$, $XC$, $XF$, $XG$, $XN$), are 98.0, 88.3 and 85.9, for Dev, and 95.7, 86.7 and 83.0, for Test, respectively.  

Table~\ref{tab:dev+test-mle-vs-analyzer}  does not include split and merge error tags as they are not present in the ALC corpus. We include all the *O {\it Other} error tags even though {\areta} does not handle them, for completeness.

\begin{table*}[t!]
    \centering{
   \includegraphics[width=0.8\textwidth]{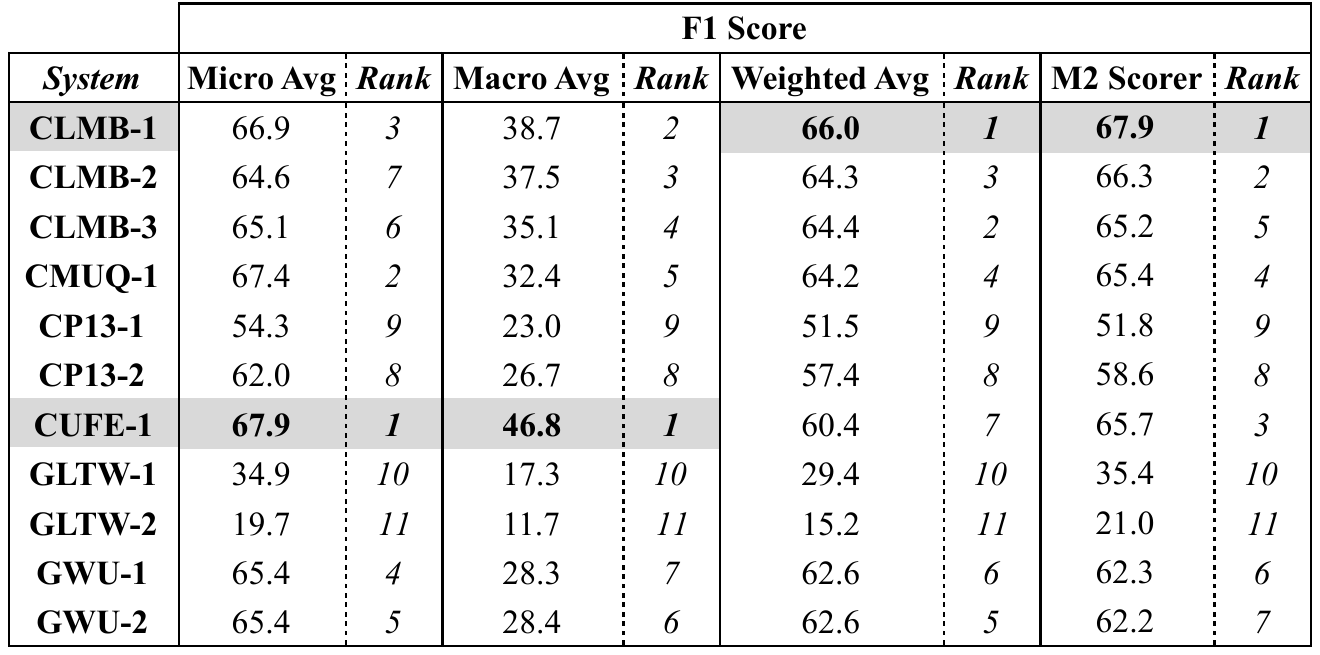}
    \caption{Comparing submitted systems from QALB 2014 shared task using M$^2$ Scorer and  F1-based metrics of {\areta}'s tags. Systems rankings are presented in italics.}
    \label{tab:framvsm2}
    }
\end{table*}

\begin{table*}[t!]
    \centering{
   \includegraphics[width=0.6\textwidth]{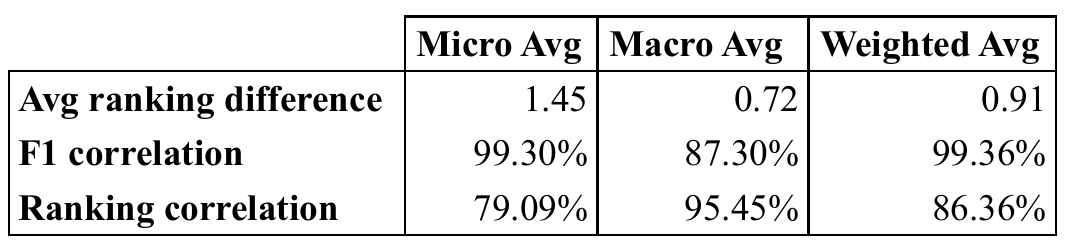}
    \caption{Correlation between the M$^2$ Scorer and {\areta}'s F1-based metrics and rankings in Table~\ref{tab:framvsm2}.}
    \label{tab:correl}
    }
\end{table*}

\subsection {Revisiting the QALB 2014 Shared Task Submissions}
\label{sec:QALB}
\paragraph{Our System vs M$^2$ Scorer}
We compare the M$^2$ Scorer results on the QALB 2014 \cite{mohit-etal-2014-first} shared task submissions  with F1-based metrics over the error tags produced by {\areta}. We calculate the {\it reference} error tags using the ($S_{raw}$, $S_{ref}$) pairs, and  compare them  to the predicted error tags using the pair ($S_{raw}$, $S_{hyp}$).  %See Section~\ref{errortype}.
The systems outputs we have access to are: CLMB \cite{rozovskaya-etal-2014-columbia}, CMUQ \cite{jeblee-etal-2014-cmuq}, CP13 \cite{tomeh-etal-2014-pipeline}, CUFE \cite{nawar-ragheb-2014-fast}, GLTW \cite{zerrouki2014autocorrection} and GWU \cite{attia-etal-2014-gwu}.

Table~\ref{tab:framvsm2} presents these results and their associated system rankings. 
Table~\ref{tab:correl} compares the F1-based metrics with the M$^2$ Scorer results presented in Table~\ref{tab:framvsm2} across all of the system outputs using Pearson correlation over F1 scores and rankings, and the average absolute ranking difference.
We observe a high correlation between the Weighted Avg and M$^2$ Scorer as well as the Micro Avg and M$^2$ Scorer. 
In terms of ranking, the Macro Avg has the highest correlation and lowest average ranking difference with M$^2$ Scorer.

According to the F1 Weighted Avg, the best performing system is CLMB-1. This matches with the M$^2$ Scorer ranking. But according to the F1 Macro Avg and Micro Avg, CUFE-1 is the best system.
We investigate the differences between these and other systems next using {\areta}'s rich error tag set.

%In addition of being able to have a high correlation with the M$^2$ Scorer, the system provides detailed error analysis of the output systems for each error type.

%\begin{table}[t!]
%    \centering{
%   \includegraphics[width=0.5\textwidth]{tables/compare_avg_f1.pdf}
%    \caption{Comparing submitted systems from QALB 2014 shared task using %M$^2$ scorer and metrics from our framework.}
%    \label{tab:framvsm2}
%    }
%\end{table}
%
%
%\begin{table}[t!]
%    \centering{
%   \includegraphics[width=0.5\textwidth]{tables/compare_ranking.pdf}
%    \caption{Systems ranking according to the different metrics.}
%    \label{tab:sys-rank}
%    }
%\end{table}

%\subsection {Detailed error analysis of submitted systems from QALB 2014 shared task}

\begin{table*}[hbt!]
    \centering{
   \includegraphics[width=1\textwidth]{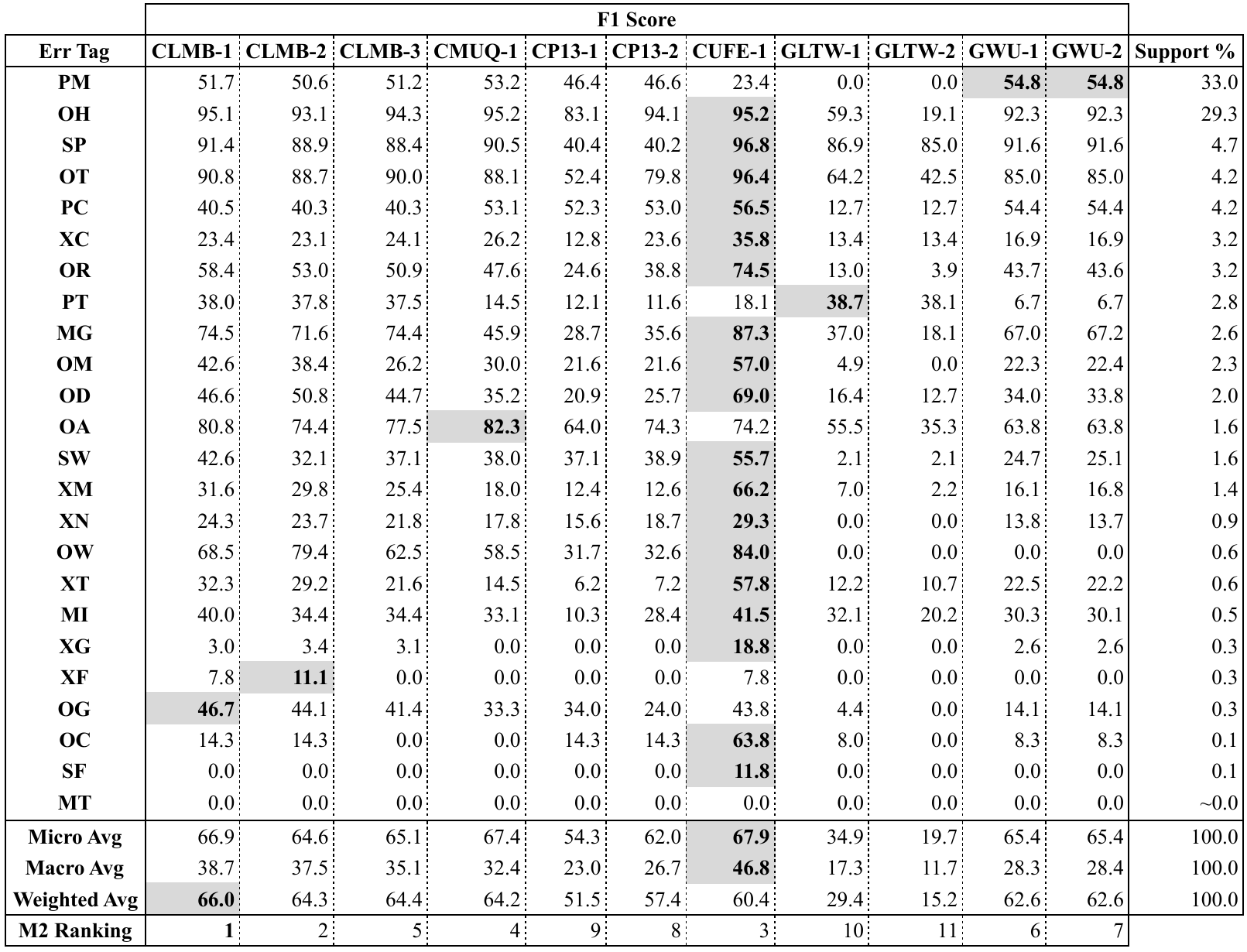}
    \caption{Comparing different system submissions on the QALB 2014 test set (classes are ordered by support).}
    \label{tab:systems-qalb-compare}
    }
\end{table*}
\paragraph{Error Type Analysis} To understand the error patterns of the QALB 2014 shared task submitted systems, we perform a detailed error analysis comparing these systems over all of the produced {\areta} error tags.
Table \ref{tab:systems-qalb-compare} presents these results ordered by support.  Only 24 of {\areta}'s 26 error tags are generated for this data set. The two missing tags are for Nun/Tanwin confusion ($ON$) and long-vowel shortening ($OS$).

We observe that the best performing system on most error types is CUFE-1 \cite{nawar-ragheb-2014-fast}. 
Interestingly, the best performing system in terms of F1 Weighted Avg is CLMB-1, but it is the best on only one minor error tag ($OG$). GWU-1 and GWU-2 are the best systems in correcting missing punctuation ($PM$), but on average they are mid-ranked. 

The $PM$ and $OH$ are the classes representing most errors from QALB 2014 test set. The results also show that most systems struggle to correct morpho-syntactic errors like gender change ($XG$) and definiteness change ($XF$). On the other hand, most systems are good at correcting orthographic errors such as Hamza ($OH$) and Ta-Marbuta ($OT$) with high accuracy.

This analysis demonstrate how {\areta} can be helpful to evaluate and diagnose errors when building Arabic GEC systems. It also motivates interesting possibilities of system combination to reach a higher performance.

%\section{Discussion}
%
%In this section, we try to address the strengths and weaknesses of the framework and discuss how it can be improved.
%
%\begin{itemize}
%    \item We used the taxonomy proposed in \cite{alfaifi2014evaluation}, and we were able to automatically match it with a good micro %F1-score (86\%) on the Al-Faifi test set. Most of the mismatches are gold errors, or sometimes different tags can be selected due to the %task's subjective nature.
%    \item We evaluated the auto vs. manual alignment from Al-Faifi, and we get 99\% correlation. This validates the quality of the aligner we %are using in the framework.
%    \item The high level of ambiguity of the Arabic language increases the computational complexity when considering the full analysis. The %mle can be used instead to make the process faster, but the quality will be a bit lower.
%\end{itemize}
%

\section{Conclusion and Future Work}

In this paper, we presented {\areta}, a publicly available automatic error type annotation system for Modern Standard Arabic targeting a modified error taxonomy based on the ALC error tagset.
We validated  {\areta}'s performance using a manually annotated blind test, where it achieved 85.8\% (Micro Avg F1 score).
We also demonstrated {\areta}'s usability in providing insightful error analyses over the submissions of the QALB 2014 shared task on Arabic text correction.

In the future, we plan to develop a new taxonomy that resolves overlapping and ambiguous error types in the ALC error tagset, and that includes more error types such as syntactic agreement and reordering operations.
We also plan to use syntactic parsers, such as \cite{Shahrour:2016:camelparser}, to model long distance dependency errors.
Naturally, we will continue to improve the various components of {\areta}, e.g., extending the handling of many-to-many word pairs, and improving specific error types.

%In this paper, we presented an automatic error annotation framework that is based on the ALC error tag taxonomy. We presented our methodology to build the suggested framework and discussed the morphological complexity of the Arabic language that increases the challenge to build such a system. Nevertheless, the framework is unsupervised and can be helpful for second language learning applications and error diagnosis of Arabic GEC systems. 

%We evaluated the quality of the automatic annotation against manually annotated data provided by \newcite{alfaifi2014evaluation} and the results show a good F1-score (weighted) 86.4\% in ALC test set. We also showed a high correlation between the metrics of the framework and the M$^2$ scorer. In addition to that, the framework can give a detailed analysis for each error type of the GEC systems, which we demonstrated on the QALB 2014 test set.  

\section*{Acknowledgments}
We would like to thank Salam Khalifa, Ossama Obeid, and Bashar Alhafni for helpful conversations and support. 

\bibliography{anthology,extra,camel-bib-v2}
\bibliographystyle{acl_natbib}

\end{document}